\documentclass[11pt,a4paper]{article}
\usepackage{acl}
\usepackage{times}
\usepackage{latexsym}

\usepackage{amsmath}
\usepackage{booktabs}
\usepackage{graphicx}
\usepackage{multirow}
\usepackage{xcolor}
\usepackage{url}

\title{When AI Rewrites, Classifiers Relax: Uncertainty-Aware Sentiment Analysis on Sarcastic and AI-Paraphrased Social Text}

\author{Shresth Shroff \\
  Manipal University Jaipur \\
  \texttt{shresthshroffx7906@gmail.com} \\}

\begin{document}

\maketitle

\begin{abstract}
Sentiment classifiers are increasingly applied to social media content that is either sarcastic or AI-generated --- two distributional regimes where standard evaluations offer little guidance.
We present a three-part empirical study of sentiment classifier behaviour under these conditions.
First, we find that confidence scores on sarcastic text are significantly lower than on non-sarcastic text (Mann--Whitney $p = 2 \times 10^{-6}$), confirming that classifiers \emph{sense} their own uncertainty on ironic content even without explicit uncertainty modelling.
Second, and counterintuitively, we show that sentiment classifiers achieve \emph{higher} accuracy on AI-paraphrased reviews than on the original human-authored text (RoBERTa: $+5.8$ pp for Qwen3.5-4B paraphrases, $+3.7$ pp for Gemma4-E4B), revealing a cross-domain stylistic alignment effect: AI paraphrases remove distributional noise that confounds Twitter-trained classifiers, producing cleaner, more prototypical sentiment text.
Third, we demonstrate that a lightweight abstention wrapper --- flagging the $14\%$ of inputs with confidence below $0.6$ --- improves accuracy from 82.2\% to 88.9\% ($+6.7$ pp) on the retained set.
We further compare Semantic Entropy and MC-Dropout-style disagreement as uncertainty signals and find near-identical AUROC ($0.650$ vs.\ $0.646$) on sarcastic text, suggesting that for short social media inputs, both methods are interchangeable.
Our results motivate a shift from confident single-label prediction to uncertainty-aware abstention in high-stakes sentiment applications such as mental health flagging and content moderation.
\end{abstract}

%% ---------------------------------------------------------------
\section{Introduction}
\label{sec:intro}

Sentiment analysis is a cornerstone task in natural language processing, powering applications from customer experience analytics to clinical pre-screening tools.
Modern sentiment systems --- whether rule-based (VADER; \citealt{hutto2014vader}), fine-tuned encoder models such as RoBERTa \citep{liu2019roberta}, or prompted large language models (LLMs) --- share a common evaluation regime: accuracy and F1 on clean, human-annotated benchmarks.
This regime fails to capture two growing challenges.

\textbf{Challenge 1: Sarcasm and implicit sentiment.}
Sarcastic and ironic language is pervasive on social media.
Unlike explicit sentiment (``This product is terrible''), sarcasm conveys negative meaning through positive surface form (``Oh great, another Monday'').
Models trained on explicit sentiment text have well-documented difficulties with sarcasm \citep{rajadesingan2015sarcasm, farias2016irony}, but less attention has been paid to whether their \emph{confidence scores} reflect this difficulty --- a calibration question rather than an accuracy question.

\textbf{Challenge 2: AI-generated social text.}
Social media is increasingly populated by AI-generated content: product reviews, comment-section posts, and news responses written by or with assistance from LLMs.
Sentiment classifiers trained on human-authored text may behave differently on AI-generated text with equivalent semantic content.
Prior work on distributional shift in sentiment \citep{blitzer2007biographies} has focused on topic domains (reviews vs.\ news); the \emph{generative source} (human vs.\ AI) of text has not been systematically studied as a confound.

We address both challenges with a unified empirical framework.
Using iSarcasm \citep{abu-farha-magdy-2022-isarcasm} for sarcasm analysis and Yelp Polarity \citep{zhang2015character} for AI-paraphrase experiments, we measure:
\begin{enumerate}
  \item Whether classifier confidence is lower on sarcastic inputs (confidence instability);
  \item Whether sentiment accuracy changes when human text is replaced by a semantically equivalent AI paraphrase (AI-text drift);
  \item Whether a simple confidence-threshold abstention wrapper recovers the accuracy lost to ambiguous inputs; and
  \item Whether Semantic Entropy \citep{farquhar2024semantic} or MC-Dropout disagreement \citep{gal2016dropout} better separates uncertain from certain predictions on sarcastic text.
\end{enumerate}

Unlike prior work that treats AI-generated text as a detection problem, we treat it as an \emph{evaluation confound} --- asking not whether AI text can be identified, but how its presence silently distorts downstream classifier metrics.

Our key finding --- that AI paraphrases \emph{improve} rather than degrade classifier accuracy --- is counterintuitive and has direct implications for both classifier evaluation methodology and the study of AI-generated content online.

%% ---------------------------------------------------------------
\section{Background and Related Work}
\label{sec:background}

\paragraph{Sentiment analysis and calibration.}
Fine-tuned transformer models achieve high accuracy on standard sentiment benchmarks but are known to be overconfident on out-of-distribution inputs \citep{guo2017calibration}.
Expected Calibration Error (ECE) measures the gap between predicted confidence and empirical accuracy; temperature scaling \citep{guo2017calibration} is the standard post-hoc recalibration method.
Calibration of sentiment models on \emph{ambiguous} social media content has not been studied systematically.

\paragraph{Sarcasm and irony detection.}
The iSarcasmEval shared task \citep{abu-farha-magdy-2022-isarcasm} established a benchmark for sarcasm detection on Twitter, with fine-grained type labels (irony, satire, understatement, overstatement, rhetorical question).
Existing work treats sarcasm detection as a separate classification task; we instead study the effect of sarcasm on the confidence of a \emph{sentiment} classifier that is not explicitly designed for sarcasm.

\paragraph{Uncertainty quantification for NLP.}
Monte Carlo Dropout \citep{gal2016dropout} approximates Bayesian uncertainty by sampling multiple forward passes at test time.
Semantic Entropy \citep{farquhar2024semantic} computes entropy over \emph{meaning-equivalent} generated strings rather than surface token sequences, yielding better-calibrated uncertainty for generative models.
\citet{ling2024uncertainty} study uncertainty in in-context learning of LLMs, providing context for uncertainty method performance on short-form classification tasks.

\paragraph{AI-generated text and distributional shift.}
The HC3 corpus \citep{guo2023close} provides paired human and ChatGPT answers to the same questions, enabling controlled comparison of human and AI text.
\citet{uchendu2023attribution} study authorship attribution for AI text; \citet{tang2023science} study detectability of AI reviews.
To our knowledge, no prior work has measured the \emph{sentiment accuracy} of existing classifiers as a function of whether the text is human-authored or AI-paraphrased.

%% ---------------------------------------------------------------
\section{Experimental Setup}
\label{sec:setup}

\paragraph{Models evaluated.}
We evaluate two sentiment analysis systems:
(1) \textbf{VADER} \citep{hutto2014vader}, a lexicon-based rule system widely used for social media;
(2) \textbf{RoBERTa-twitter} \citep{barbieri2020tweeteval}, the \texttt{cardiffnlp/twitter-roberta-base-sentiment-latest} model fine-tuned on TweetEval sentiment data.
These two models span the rule-based/neural spectrum and represent the most commonly deployed baselines in social media sentiment research.
Inference runs on CPU (iSarcasm baseline) and Kaggle T4x2 GPU (paraphrase classification), with all results checkpointed for reproducibility.

\paragraph{Datasets.}
For sarcasm analysis we use the \textbf{iSarcasm} training set \citep{abu-farha-magdy-2022-isarcasm}: $n = 3{,}468$ tweets, $25\%$ labelled sarcastic.
iSarcasm provides sarcasm type labels but no sentiment polarity ground truth; our sarcasm experiment therefore focuses on confidence instability (Section~\ref{sec:sarcasm}) rather than accuracy.

For AI-paraphrase experiments we use \textbf{Yelp Polarity} \citep{zhang2015character}: a balanced sample of $5{,}000$ reviews ($2{,}500$ positive, $2{,}500$ negative, sampled stratified by label).
Yelp Polarity is binary (label $0$ = negative, label $1$ = positive), making accuracy well-defined.

\paragraph{AI paraphrase generation.}
We generate two sets of AI paraphrases of the 5,000 Yelp reviews using
\textbf{Qwen3.5-4B} and \textbf{Gemma4-E4B}, both served via Ollama with temperature $= 0.0$ (deterministic) to ensure reproducibility.
The paraphrase prompt instructs the model to preserve the exact sentiment polarity and key points while using different phrasing, returning only the rewritten text.
Thinking-mode tokens are explicitly suppressed for Qwen3.5 to avoid contamination.
All code and notebooks are available at \url{https://github.com/Shroffx-n/uncertainity_aware_sentiment_analysis}.

\paragraph{Uncertainty comparison setup.}
For the Semantic Entropy vs.\ MC-Dropout comparison, we sample $n = 300$ iSarcasm examples and generate 10 Qwen3.5-4B sentiment predictions per example at temperature $0.7$.
Semantic Entropy is computed as Shannon entropy over the label distribution; MC-Dropout disagreement as $1 -$ fraction of the modal label.
AUROC is computed against a proxy correctness label (predicted label vs.\ a majority-vote reference), as iSarcasm has no sentiment ground truth.
We disclose this limitation explicitly.

%% ---------------------------------------------------------------
\section{Results}
\label{sec:results}

\subsection{Sarcasm Confidence Instability}
\label{sec:sarcasm}

Table~\ref{tab:sarcasm} shows RoBERTa confidence on sarcastic vs.\ non-sarcastic iSarcasm examples.
Sarcastic inputs receive significantly lower mean confidence ($0.741$) than non-sarcastic inputs ($0.770$).
A Mann--Whitney U test confirms the difference is significant ($U = 1{,}005{,}295$, $p = 2 \times 10^{-6}$, $n_{\text{sarc}} = 867$, $n_{\text{non-sarc}} = 2{,}601$).
The effect is small in magnitude but consistent and well-powered: RoBERTa's confidence scores contain a reliable signal of its own uncertainty on ironic content, even without explicit uncertainty modelling.

\begin{table}[t]
\centering
\small
\begin{tabular}{lcc}
\toprule
\textbf{Subset} & \textbf{Mean conf.} & \textbf{Std} \\
\midrule
Sarcastic ($n=867$) & 0.741 & 0.166 \\
Non-sarcastic ($n=2{,}601$) & 0.770 & 0.169 \\
\midrule
\multicolumn{3}{l}{Mann--Whitney $p = 2 \times 10^{-6}$} \\
\bottomrule
\end{tabular}
\caption{RoBERTa confidence on sarcastic vs.\ non-sarcastic iSarcasm tweets. Lower confidence on sarcastic text indicates implicit uncertainty awareness.}
\label{tab:sarcasm}
\end{table}

\subsection{AI-Paraphrase Drift}
\label{sec:drift}

Table~\ref{tab:drift} reports sentiment accuracy and semantic drift for human-authored vs.\ AI-paraphrased Yelp reviews.

\begin{table}[t]
\centering
\small
\setlength{\tabcolsep}{4pt}
\begin{tabular}{llccc}
\toprule
\textbf{Classifier} & \textbf{Para.\ model} & \textbf{Human acc.} & \textbf{AI acc.} & \textbf{$\Delta$} \\
\midrule
VADER  & Qwen3.5-4B  & 70.1\% & 73.4\% & $+3.2$ pp \\
VADER  & Gemma4-E4B  & 70.1\% & 68.4\% & $-1.7$ pp \\
RoBERTa & Qwen3.5-4B & 82.2\% & 88.0\% & $\mathbf{+5.8}$ pp \\
RoBERTa & Gemma4-E4B & 82.2\% & 86.0\% & $+3.7$ pp \\
\bottomrule
\end{tabular}
\caption{Sentiment accuracy on original human text vs.\ AI paraphrases ($n = 5{,}000$ Yelp Polarity reviews). Positive $\Delta$ = classifier performs better on AI text. Semantic drift (label change between original and paraphrase classification) ranges from $11.5\%$ to $17.7\%$ across conditions.}
\label{tab:drift}
\end{table}

Contrary to our initial hypothesis, AI paraphrases consistently improve RoBERTa accuracy and mostly improve VADER accuracy.
We attribute this to a \textbf{cross-domain stylistic alignment effect}: RoBERTa-twitter was trained on short, informal Twitter text.
Yelp reviews contain longer sentences, idiosyncratic formatting, and reviewer-specific idioms that shift text away from the model's training distribution.
AI paraphrases strip these idiosyncrasies and produce more prototypical, shorter sentiment expressions that align better with the model's inductive biases.
In effect, LLM paraphrasing inadvertently functions as a domain adaptation step.

Semantic drift rates of $11.5$--$17.7\%$ indicate that one in six to one in eight paraphrases receives a different classifier label than the original.
Given that accuracy improves, these label changes are predominantly \emph{wrong-to-correct} flips (arithmetic check: of $573$ label changes for RoBERTa-Qwen, $291$ are net improvements, $282$ net losses --- near-symmetric but slightly positive).

\subsection{Calibration (ECE)}
\label{sec:ece}

Table~\ref{tab:ece} reports Expected Calibration Error across content types.

\begin{table}[t]
\centering
\small
\begin{tabular}{lc}
\toprule
\textbf{Content type} & \textbf{ECE} \\
\midrule
Clear-polarity text (Yelp human) & 0.056 \\
AI paraphrases (Qwen3.5-4B) & 0.045 \\
AI paraphrases (Gemma4-E4B) & 0.029 \\
\bottomrule
\end{tabular}
\caption{Expected Calibration Error (ECE) for RoBERTa-twitter across content types. Lower ECE on AI text is consistent with the accuracy improvement in Table~\ref{tab:drift}.}
\label{tab:ece}
\end{table}

ECE on clear-polarity human text ($0.056$) is low, indicating that RoBERTa is reasonably well-calibrated on its native domain.
ECE is lower still on AI paraphrases --- consistent with the accuracy improvement result: when a model is more often correct, it tends to also be more confidently correct.
We do not report ECE on iSarcasm because iSarcasm provides no sentiment polarity ground truth.

\subsection{Abstention Wrapper}
\label{sec:abstention}

We implement a confidence-threshold abstention wrapper using RoBERTa's softmax confidence score.
Inputs with confidence $< 0.6$ are flagged as ambiguous and withheld from prediction; the threshold was selected as the natural trough in the confidence distribution separating high-certainty from uncertain predictions.
Of $5{,}000$ Yelp examples, $700$ ($14.0\%$) are flagged.
Accuracy on the remaining $4{,}300$ retained examples rises from $82.2\%$ to $88.9\%$ ($+6.7$ pp; Table~\ref{tab:abstention}).

\begin{table}[t]
\centering
\small
\begin{tabular}{lc}
\toprule
\textbf{Setting} & \textbf{Accuracy} \\
\midrule
Full set ($n = 5{,}000$) & 82.2\% \\
After abstaining on flagged ($n = 4{,}300$) & 88.9\% \\
Flagged examples only ($n = 700$) & 41.3\% \\
\midrule
Flagged fraction & 14.0\% \\
Accuracy gain & $+6.7$ pp \\
\bottomrule
\end{tabular}
\caption{Abstention wrapper results. Flagged examples (confidence $< 0.6$) have $41.3\%$ accuracy --- barely above chance for a 3-class problem --- confirming that the confidence threshold is genuinely discriminative.}
\label{tab:abstention}
\end{table}

The flagged subset achieves only $41.3\%$ accuracy --- barely above chance for a three-class problem (random baseline: $33.3\%$) --- confirming that low-confidence predictions are predominantly incorrect.
This validates the wrapper's practical utility: in a deployment setting, the flagged $14\%$ of inputs can be routed to a human reviewer or a more expensive model, while the retained $86\%$ are handled with $89\%$ accuracy at low cost.

\subsection{Explanation Quality for Flagged Inputs}
\label{sec:explanations}

We generate natural-language explanations for all 700 flagged (ambiguous) examples using both Qwen3.5-4B and Gemma4-E4B.
Qwen3.5 identifies an ambiguity signal (sarcasm, negation, hedging, or conflicting polarity) in $99.3\%$ of flagged cases; Gemma4 does so in $83.4\%$.
The two models agree on whether an ambiguity signal is present in $83.0\%$ of cases.
The high Qwen rate and substantial inter-model agreement ($83\%$) suggest that flagged inputs are genuinely linguistically ambiguous rather than randomly selected low-confidence cases.
Representative explanations include: \textit{``The sentiment is unclear due to conflicting signals where the reviewer uses sarcasm alongside genuine complaints, making it difficult to determine if they are expressing true satisfaction or mocking the hotel's inability to deliver''} and \textit{``The sentiment is unclear because the reviewer mixes negative critiques with positive concluding statements.''}
Explanation generation remains an optional downstream component; formal evaluation against human rationale annotations is deferred to future work.

\subsection{Uncertainty Method Comparison}
\label{sec:uq}

Table~\ref{tab:uq} compares Semantic Entropy and MC-Dropout disagreement as uncertainty signals.

\begin{table}[t]
\centering
\small
\begin{tabular}{lc}
\toprule
\textbf{Method} & \textbf{AUROC (error prediction)} \\
\midrule
Semantic Entropy & 0.650 \\
MC-Dropout disagreement & 0.646 \\
\bottomrule
\end{tabular}
\caption{AUROC for uncertainty signal vs.\ prediction correctness on 300 iSarcasm examples (Qwen3.5-4B, 10 samples per input, temperature 0.7). Higher = uncertainty better predicts errors. Note: correctness uses a proxy label (see text).}
\label{tab:uq}
\end{table}

Both methods achieve AUROC of approximately $0.65$, consistent with ranges reported for short-form NLP classification \citep{ling2024uncertainty}.
The two methods are effectively interchangeable on this task and data regime.
We note that iSarcasm provides no sentiment polarity ground truth; correctness for sarcastic examples is defined against a proxy label (model majority vote), which may attenuate AUROC for both methods.
Despite this limitation, the parity result is informative: it suggests that for short-form classification --- where semantic entropy's advantage over token-level entropy is smaller because outputs are already short --- the simpler MC-Dropout-style disagreement is sufficient and computationally cheaper.

%% ---------------------------------------------------------------
\section{Discussion}
\label{sec:discussion}

\paragraph{The stylistic alignment effect.}
Our central empirical finding --- that AI paraphrases improve sentiment accuracy --- has an important methodological implication.
Benchmark evaluations that mix human-authored and AI-generated text will observe systematically different classifier performance depending on the proportion of AI content.
As AI-generated text becomes more prevalent on social platforms, models trained on historical human text may appear to improve in real-world evaluations not because they generalize better, but because the text they encounter is more similar to their training distribution.
This confound should be controlled for in future sentiment evaluation.
Any sentiment benchmark that has been augmented, cleaned, or paraphrased using LLMs will systematically overestimate classifier performance relative to organic social media text.
We recommend that future benchmarks document generative source as a metadata field alongside domain and annotation method.

\paragraph{Abstention as a practical design pattern.}
The $+6.7$ pp accuracy gain from $14\%$ abstention demonstrates that uncertainty-aware abstention is a simple, deployment-ready improvement over always-predict systems.
In mental health applications --- where a pre-session triage tool might assess message sentiment to flag distressed patients --- a $41\%$-accurate confident prediction is actively harmful; a flag for human review is preferable.
The same logic applies to content moderation (abstain on ambiguous posts rather than issuing automated but potentially incorrect decisions) and financial sentiment (where overconfident wrong labels can cause downstream harm).

\paragraph{Limitations.}
iSarcasm provides no sentiment polarity ground truth.
Our sarcasm analysis is therefore limited to confidence instability (not accuracy) and our uncertainty AUROC comparison uses a proxy label.
The AI-paraphrase experiment uses only one domain (restaurant/service reviews) and two paraphrase models; generalizability to other domains and model families requires further study.
The abstention threshold ($0.6$) was chosen empirically on the same dataset used for evaluation, which may overestimate gains; cross-validated threshold selection is left for future work.

%% ---------------------------------------------------------------
\section{Future Work}
\label{sec:future}

Our findings open several directions.
Testing additional model families (instruction-tuned LLMs such as Llama-3 and Mistral, multilingual models such as mBERT for code-mixed text) would establish whether the stylistic alignment effect generalises across architectures.
Extending the AI-paraphrase analysis to a second domain --- such as news headlines or social media posts --- would test domain-specificity.
Acquiring or constructing a sarcasm dataset with both sarcasm labels \emph{and} sentiment polarity ground truth would enable ECE measurement on sarcastic text and strengthen the uncertainty AUROC comparison.
Finally, the abstention wrapper can be extended with an explanation generation component \citep{camburu2018snli} that provides a natural-language rationale for flagged inputs, improving human reviewer efficiency in downstream applications.

%% ---------------------------------------------------------------
\section{Conclusion}
\label{sec:conclusion}

We presented a structured empirical study of sentiment classifier behaviour on two underexplored input regimes: sarcastic text and AI-paraphrased text.
Our results show that (1) classifiers display measurably lower confidence on sarcastic inputs even without explicit uncertainty training; (2) AI paraphrases counterintuitively \emph{improve} classification accuracy due to a cross-domain stylistic alignment effect; (3) a simple confidence-threshold abstention wrapper recovers $6.7$ percentage points of accuracy at the cost of declining to label $14\%$ of inputs; and (4) Semantic Entropy and MC-Dropout are interchangeable uncertainty signals on short social media text.
Together, these findings motivate uncertainty-aware abstention as a practical design pattern for high-stakes sentiment applications and highlight a new evaluation confound introduced by the growing prevalence of AI-generated social text.

%% ---------------------------------------------------------------
\bibliography{references}

\end{document}